\documentclass[10pt,twocolumn,letterpaper]{article}

\usepackage[pagenumbers]{cvpr} % To force page numbers, e.g. for an arXiv version

\usepackage{graphicx}
\usepackage{amsmath}
\usepackage{amssymb}
\usepackage{booktabs}
\usepackage{bm}
\usepackage{comment}  % added for inserting some comments.
\usepackage{pifont}% http://ctan.org/pkg/pifont
\usepackage[pagebackref,breaklinks,colorlinks]{hyperref}
\usepackage{multirow}

\usepackage[capitalize]{cleveref}
\crefname{section}{Sec.}{Secs.}
\Crefname{section}{Section}{Sections}
\Crefname{table}{Table}{Tables}
\crefname{table}{Tab.}{Tabs.}

\def\cvprPaperID{8161} % *** Enter the CVPR Paper ID here
\def\confName{CVPR}
\def\confYear{2024}

\begin{document}

\title{NeISF: Neural Incident Stokes Field for Geometry and Material Estimation}

\author{Chenhao Li$^{1,3}$,\quad Taishi Ono$^2$,\quad Takeshi Uemori$^1$,\quad Hajime Mihara$^1$, \\ Alexander Gatto$^2$,\quad Hajime Nagahara$^3$,\quad Yusuke Moriuchi$^1$\\
\\
{Sony Semiconductor Solutions Corporation$^1$,\quad Sony Europe B.V.$^2$,\quad  Osaka University$^3$}
%\and
%Sony Group Corporation\\
}
\twocolumn[{
        \maketitle
        \vspace{-3.5em}
	\begin{center}
            \captionsetup{type=figure}
		\includegraphics[width=\textwidth]{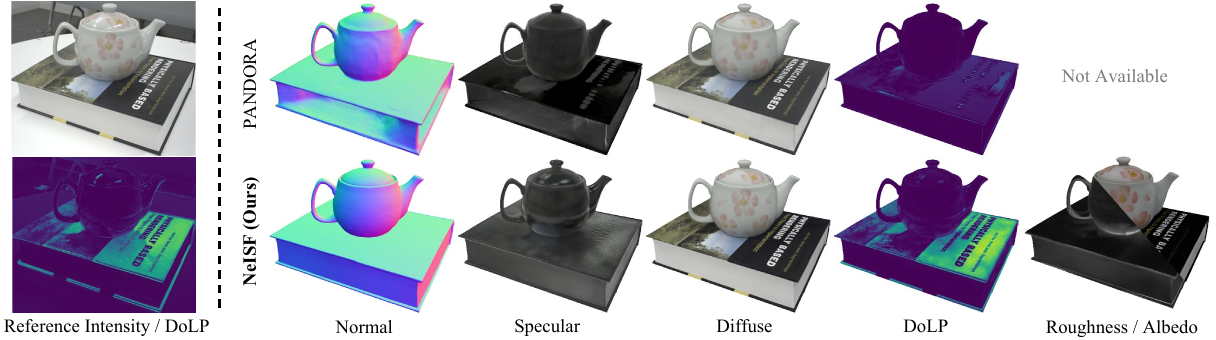}
		\captionof{figure}{NeISF reconstructs highly accurate shapes and materials using polarization cues. The inter-reflection between the teapot and the book is clearly observed in our specular intensity, while PANDORA \cite{dave2022pandora} is heavily affected by the textures and does not correctly reconstruct the inter-reflection because it only assumes single-bounced illumination. DoLP denotes the degree of linear polarization.}		
	\label{fig:compare_sakurapot}
        \end{center}
}]

%%%%%%%%% ABSTRACT
\begin{abstract}
% What is muti-view inverse rendering
Multi-view inverse rendering is the problem of estimating the scene parameters such as shapes, materials, or illuminations from a sequence of images captured under different viewpoints.
% the limitation of single-bounce methods
% examples: NeRD, PANDORA
Many approaches, however, assume single light bounce and thus fail to recover challenging scenarios like inter-reflections.
% Extending those single-bounce methods to multi-bounce is not easy (requires further constraints)
% examples: NeILF++ uses so many assumptions 
On the other hand, simply extending those methods to consider multi-bounced light requires more assumptions to alleviate the ambiguity.
% We propose NeISF
To address this problem, we propose Neural Incident Stokes Fields (NeISF), a multi-view inverse rendering framework that reduces ambiguities using polarization cues.
% The motivation for using polarization cues
The primary motivation for using polarization cues is that it is the accumulation of multi-bounced light, providing rich information about geometry and material.
% our novelties
Based on this knowledge, the proposed incident Stokes field efficiently models the accumulated polarization effect with the aid of an original physically-based differentiable polarimetric renderer.
% We achieve SoTA for real and synthetic datasets with these two novelties 
Lastly, experimental results show that our method outperforms the existing works in synthetic and real scenarios.
\end{abstract}

\begin{figure*}[t]
  \centering
  \includegraphics[width=\linewidth]{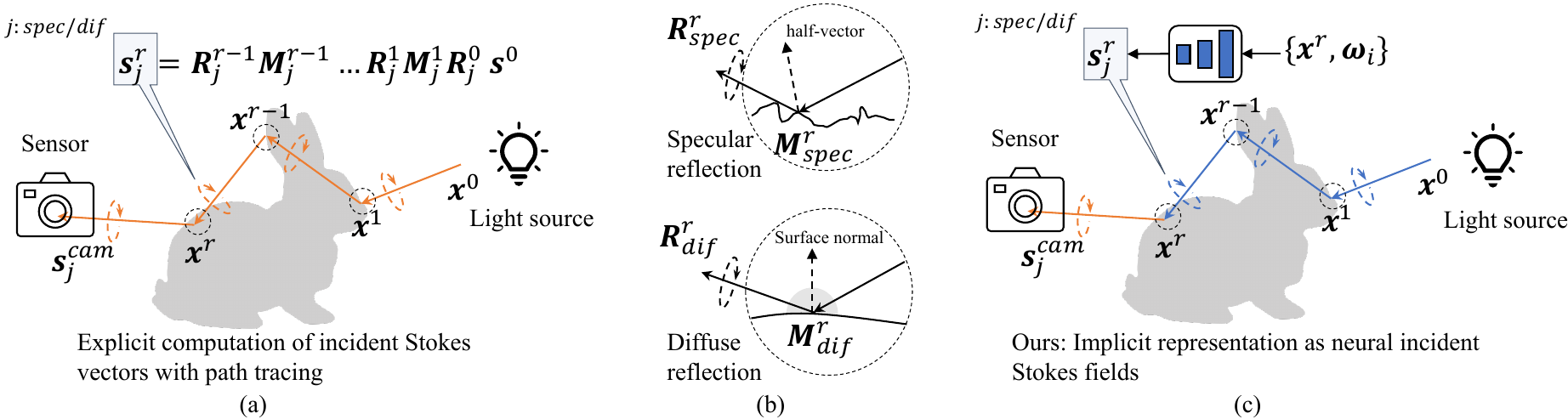}
  \caption{Concept of our incident Stokes fields. The orange paths are explicitly computed, while the blue paths are implicitly represented. (a) In the traditional path tracer, the incident Stokes vectors are computed by the recursive multiplication of Stokes vectors, rotation Mueller matrices, and pBRDF Mueller matrices. (b) The diffuse and specular pBRDF matrices have different reference frames. Thus, the rotation matrices should be treated separately. (c) Given the positions of the interaction points and the directions of the incident light, we use MLPs to implicitly record the already-rotated incident Stokes vectors of diffuse and specular components, separately.}
  \label{fig:concept}
\end{figure*}

%%%%%%%%% BODY TEXT
\section{Introduction}

% What is inverse rendering, main challenge, and two groups of solutions
Inverse rendering aims at decomposing the target scene into parameters such as geometry, material, and lighting. It is a long-standing task for computer vision and computer graphics and has many downstream tasks, such as relighting, material editing, and novel-view synthesis. The main challenge of inverse rendering is that many combinations of the scene parameters can express the same appearance, called the ambiguity problem. Solutions to the ambiguity can be roughly divided into two categories: simplifying the scene and utilizing more information. For the first group, some studies assume a Lambertian Bidirectional Reflectance Distribution Function (BRDF) \cite{sengupta2018sfsnet}, a single point light source \cite{deschaintre2018single}, or a near-planer geometry \cite{li2018materials}. For the second group, additional information such as multiple viewpoints \cite{kim2016multi}, additional illuminations \cite{Boss2020-TwoShotShapeAndBrdf}, multi-spectral images \cite{li2021spectral}, depth information \cite{ku2022differentiable}, and polarization cues \cite{zhao2022polarimetric} has been extensively explored. 

% implicit representations (NeRF), and the limitation
Most of the aforementioned methods use explicit representations of scene parameters. 
On the other hand, Neural Radiance Fields (NeRF) \cite{mildenhall2020nerf} shows the successful use of implicit representations. 
Although NeRF achieves remarkable performance on novel-view synthesis, it does not decompose the scene into the parameters.
Thus, many approaches try to extend the NeRF representation to solve inverse rendering, and solutions to the ambiguity problem can still be categorized into the two groups stated above.
% Solutions to the ambiguity problem still follow two groups
For the first group, assumptions such as a smooth roughness field \cite{yao2022neilf}, low roughness \cite{tiwary2023orca}, known lighting \cite{srinivasan2021nerv, yang2022s3nerf}, single light bounce \cite{boss2021nerd, boss2022-samurai}, collocated flashlight \cite{zhang2022iron}, or a Lambertian surface \cite{zhang2023neilf++} are proposed to stabilize the training. However, these assumptions severely limit the scope of target scenes.
For the second group, various types of cues such as depth images \cite{attal2021torf}, azimuth maps \cite{mvas2023cao}, multiple lights \cite{li2022neural, ling2023shadowneus} and multi-spectral information \cite{poggi2022xnerf} are investigated. Additionally, polarization is also examined in this group. To our knowledge, PANDORA \cite{dave2022pandora} first combines the implicit representations and polarization cues for the diffuse-specular reflection separation and the geometry estimation. However, due to the entangled representation of the incident light and surface reflectance, BRDF parameters are not estimated. In addition, they assume a single light bounce and unpolarized incident light. These limitations lead to our key research question: Can polarization cues disambiguate the full NeRF-based inverse rendering?

% our proposal
We propose Neural Incident Stokes Fields (NeISF), an inverse rendering method using polarization cues and implicit representations. It takes multi-view polarized images of a static object with known object masks and camera poses but with unknown geometry, material, and lighting. Based on the implicit representation for the multi-bounced light \cite{yao2022neilf, zhang2023neilf++}, the proposed incident Stokes fields effectively extend this representation to include the polarization cues.
Specifically, instead of explicitly modeling every single light bounce as shown in Fig. \ref{fig:concept} (a), we use coordinate-based multi-layer perceptrons (MLPs) to record Stokes vectors of all the second-last bounces (Fig. \ref{fig:concept} (c)). After that, we introduce a physically-based polarimetric renderer to compute Stokes vectors of the last bounces using a polarimetric BRDF model proposed by Baek \etal \cite{baek2018simultaneous} (Baek pBRDF). The challenging part of extending an unpolarized incident light field to a polarized one is that we must ensure that the Stokes vectors are properly rotated to share the same reference frame with the Mueller matrices.
Furthermore, the diffuse and specular components have different reference frames, which makes the problem more complicated. This is because the reference frame of diffuse Mueller matrices depends on the surface normal, while the reference frame of specular Mueller matrices depends on the microfacet normal (Fig. \ref{fig:concept} (b)). To solve this issue, we propose to implicitly record the rotation of the second-last bounce for the diffuse and specular components separately. More specifically, given the position and direction of the incident light, we use MLPs to record the already rotated Stokes vectors of diffuse and specular components independently. Our light representation is capable of handling challenging scenes including those that have inter-reflections. In addition, the polarization cues can provide a wealth of information on geometry, material, and light, making it easier to solve inverse rendering compared to unpolarized methods. To comprehensively evaluate the proposed approach, we construct two polarimetric HDR datasets: a synthetic dataset rendered by Mitsuba 3.0 \cite{Jakob2020DrJit}, and a real-world dataset captured by a polarization camera. Fig. \ref{fig:compare_sakurapot} shows that our model outperforms the existing methods. Our contributions are summarized as follows:
\begin{itemize}
  \item This method introduces a unique representation, which implicitly models multi-bounce polarized light paths with the rotation of Stokes vectors taken into account.
  \item To perfectly integrate the representation into the training pipeline, we introduce a differentiable physically-based polarimetric renderer.
  \item Our method achieves state-of-the-art performance on both synthetic and real scenarios.
  \item Our real and synthetic multi-view polarimetric datasets and implementation are publicly available.
\end{itemize}

\section{Related Works}
\noindent\textbf{Inverse Rendering}
We roughly divide the existing inverse rendering works into two groups, which are learning-based and optimization-based methods. Most of the learning-based inverse rendering works \cite{li2023inverse, li2018learning, sengupta2019neural, li2020inverse, Wang_2021_ICCV, zhu2022irisformer, li2022phyir, boss2020two, sang2020single, lichy2021shape} are single-view approaches. They mainly rely on large-scale synthetic training datasets because acquiring the ground truth material, geometry, and lighting parameters is labor-intensive and time-consuming. A well-known problem of using synthetic training data is the domain gap, where the trained model often fails to give reasonable results for the real-world scene. Optimization-based methods \cite{mukaigawa2010analysis, nishino2009directional, barron2014shape, yoon2010joint, lombardi2015reflectance, oxholm2014multiview, lombardi2016radiometric}, also known as analysis by synthesis, are the other direction to solve inverse rendering. The recent breakthrough of optimization-based methods is dominated by the differentiable rendering \cite{nicolet2023recursive, li2018differentiable, Zhang:2019:DTRT, Zhang:2020:PSDR}. Differentiable renderers like Mitsuba \cite{nimier2019mitsuba} are able to backpropagate the gradients to physical parameters even when the light is bounced multiple times. However, it requires a huge computational cost and memory consumption when handling a complex scene. NeRF-based inverse rendering can also be classified as the optimization-based method. Compared to the explicit representation of scene parameters, the compactness and effectiveness of neural implicit representation have been verified. Details will be introduced in the next part.

\noindent\textbf{Neural Implicit Fields}
NeRF \cite{mildenhall2020nerf} achieved photorealistic performance for novel-view synthesis utilizing the effectiveness of implicit neural representation. However, inverse rendering is not directly supported due to the entangled representation. This limitation has opened up a new research field on neural implicit fields-based inverse rendering.

Some works only focus on geometry estimation. Representative works such as IDR \cite{yariv2020multiview}, NeuS \cite{wang2021neus}, VolSDF \cite{yariv2021volume}, and BakedSDF \cite{yariv2023bakedsdf} can be classified into this group. They disentangle the geometry but use an entangled representation of the lighting and material. Attempts to complete the disentanglement have been widely studied. Early works only consider the direct lighting represented by a spherical Gaussian \cite{boss2021nerd, zhang2021physg}, an environment map \cite{zhang2021nerfactor}, or split-sum approximation \cite{boss2021neuralpil, munkberg2022extracting}. These direct lighting-based works are not capable of handling complex effects like inter-reflection. Later, several works \cite{wang2023neural, li2023multi, zhang2022modeling, jin2023tensoir, wu2023nefii, zhu2023i2} that also consider indirect lighting have been reported. One simple but efficient solution is Neural Radiosity \cite{hadadan2021neural}, which records a part of light bounces using MLPs. Inspired by them, many works \cite{yao2022neilf, zhang2023neilf++, hadadan2023inverse} also use such kind of light representation for inverse rendering. We extend their idea by proposing the neural incident Stokes fields to model the multi-bounced polarimetric light propagation.

\noindent\textbf{Polarization}
Polarization is one of the properties of electromagnetic waves that specifies the geometrical orientation of the oscillations. An important phenomenon of polarization is that it changes after interacting with objects, providing rich information for a variety of applications including inverse rendering. Since the release of commercial polarization cameras \cite{yamazaki2016four}, it has become easier to capture polarized images, and polarization research has become more active. Various applications such as the estimation of shape \cite{ngo2015shape, kondo2020accurate, fukao2021polarimetric, ichikawa2021shape, ba2020deep, lei2022shape, shao2022transparent, chen2022perspective, kadambi2015polarized, zou20203d}, material \cite{azinovic2023high, deschaintre2021deep, hwang2022sparse, baek2022all, duan2023end}, pose \cite{cui2019polarimetric,zou2022human, gao2022polarimetric}, white balance \cite{ono2022degree}, reflection removal \cite{lei2020polarized, lyu2019reflection, li2020reflection}, segmentation \cite{kalra2020deep, mei2022glass, liang2022multimodal}, and sensor design \cite{kurita2023simultaneous} have been explored. 

So far, attempts to combine NeRF and polarization mainly focused on extending the intensity fields to the polarimetric (pCON \cite{peters2023pcon}) fields or Spectro-polarimetric (NeSpoF \cite{kim2023neural}) fields for novel-view synthesis. Namely, they do not use polarization for inverse rendering. PANDORA \cite{dave2022pandora} is the first work that combines polarization cues and NeRF for inverse rendering purposes. They train coordinate-based MLPs to estimate normals, diffuse radiance, and specular radiance. After that, the estimated normals, diffuse radiance, and specular radiance are combined by a simplified renderer to generate the outgoing Stokes vectors. The main limitation of PANDORA can be considered as follows. First, it does not support the inverse rendering of BRDF parameters. Because the diffuse and specular radiance entangles the incident light, BRDF, and normals. Second, they assume an unpolarized incident light. This violates the common situation in the real world where the light has already bounced and become polarized before hitting the object. In contrast, the rendering process of our method is physically based, making it possible to fully disentangle the material, geometry, and lighting. Additionally, we do not require an unpolarized incident light assumption.

\section{Preliminary}
We briefly introduce the mathematics used to describe the polarimetric light propagation, BRDF, and rendering equation. Please refer to the supplementary material for the detailed version.

\subsection{Stokes-Mueller multiplication}
The polarization state of the light can be represented as a Stokes vector $\mathbf{s} \in \mathbb{R} ^ 3$. It has three elements $[s_0, s_1, s_2]$, where $s_0$ is the unpolarized light intensity, $s_1$ is the $0^{\circ}$ over $90^{\circ}$ linear polarization, and $s_2$ is the $45^{\circ}$ over $135^{\circ}$ linear polarization.
We do not consider the fourth dimension representing circular polarization in this paper. The light-object interaction can be expressed by the multiplication of Stokes vectors and Mueller matrices:
\begin{equation}
    \label{equ:mueller-rotation-stokes}
    \mathbf{s}^\text{out} = \mathbf{M} \cdot \mathbf{R} \cdot \mathbf{s}^\text{in},
\end{equation}
where $\mathbf{M} \in \mathbb{R} ^ {3 \times 3}$ is the Mueller matrix representing the optical property of the interaction point, $\mathbf{s}^\text{in}$ and $\mathbf{s}^\text{out}$ are the incident and outgoing Stokes vectors. $\mathbf{R} \in \mathbb{R} ^ {3 \times 3}$ is the rotation matrix which depends on the relative angle of the reference frames of $\mathbf{s}^\text{in}$ and $\mathbf{M}$. It must also be multiplied, as the Stokes-Mueller multiplication is only valid when they share the same reference frame.

\subsection{Polarimetric BRDF}
In Baek pBRDF, the diffuse and specular components are modeled separately. The diffuse component $\mathbf{M}_\text{dif}$ describes the process of transmitting from the outside to inside, subsurface scattering, and transmitting from the inside to outside. It can be formulated as follows:

\begin{equation}
    \mathbf{M}_\text{dif} = (\frac{\bm{\rho}}{\pi}\cos{\theta_i}) \mathbf{F}^T_o \cdot \mathbf{D} \cdot \mathbf{F}^T_i.
\end{equation}
$\bm{\rho} \in \mathbb{R}^3$ is the diffuse albedo, $\theta_{i,o}$ denotes the incident / outgoing angle, $\mathbf{D} \in \mathbb{R}^{3 \times 3}$ is a depolarizer, and $\mathbf{F}^T_{i,o} \in \mathbb{R}^{3 \times 3}$ is the Fresnel transmission term.
The specular component describes the microfacet surface reflection:
\begin{equation}
    \mathbf{M}_\text{spec} = \mathbf{k_s} \frac{DG}{4\cos{\theta_o}} \mathbf{F}^R,
\end{equation}
where $\mathbf{k_s} \in \mathbb{R}^3$ is the specular coefficient, $D$ is the GGX distribution function \cite{walter2007microfacet},  $G$ is the Smith function, and $\mathbf{F}^R \in \mathbb{R}^{3 \times 3}$ is the Fresnel reflection.

\subsection{Polarimetric rendering equation}

According to Eq. \ref{equ:mueller-rotation-stokes}, we can obtain the polarimetric version of the Rendering Equation \cite{kajiya1986rendering}:

\begin{equation}
    \label{equ:rendering equation}
    \mathbf{s}^\text{cam} = \mathbf{R}^\text{cam} \cdot \int_\Omega  \mathbf{M} \cdot \mathbf{R}^\text{in} \cdot \mathbf{s}^\text{in} \,d \bm{\omega_i},
\end{equation}
where $\mathbf{s}^\text{cam}$ is the Stokes vector captured by the camera, $\mathbf{R}^\text{cam}$ is the rotation matrix from the Mueller matrix to the camera's reference frame, rotation matrix $ \mathbf{R}^\text{in}$ rotates the incident Stokes vector $\mathbf{s}^\text{in}$ to the reference frame of Mueller matrix $\mathbf{M}$, $\bm{\omega_i} \in \mathbb{R}^3$ is the incident direction. 

Furthermore, Baek pBRDF handles the rotation matrices of diffuse and specular components in a different manner.
Because the reference frame of the diffuse Mueller matrix $\mathbf{M}_\text{dif}$ depends on the surface normal, while the reference frame of the specular Mueller matrix $\mathbf{M}_\text{spec}$ depends on the microfacet normal (halfway vector) as shown in (Fig. \ref{fig:concept} (b)). Thus, for the diffuse component, the Eq. \ref{equ:rendering equation} should be rewritten as:
\begin{equation}
    \label{equ:rendering equation dif}
    \mathbf{s}^\text{cam}_\text{dif} = \mathbf{R}^\text{cam}_\text{dif}  \cdot \int_\Omega  \mathbf{M}_\text{dif}  \cdot \mathbf{R}^\text{in}_\text{dif}  \cdot \mathbf{s}^\text{in} \,d \bm{\omega_i},
\end{equation}
and for the specular component:
\begin{equation}
    \label{equ:rendering equation spec}
    \mathbf{s}^\text{cam}_\text{spec} =  \int_\Omega  \mathbf{R}^\text{cam}_\text{spec}  \cdot \mathbf{M}_\text{spec}  \cdot \mathbf{R}^\text{in}_\text{spec}  \cdot \mathbf{s}^\text{in} \,d \bm{\omega_i}.
\end{equation}
Note that for the specular component, $\mathbf{R}^\text{cam}_\text{spec}  \cdot$ should be placed into the integral, as the microfacet normal changes according to the incident direction $\bm{\omega_i}$.

\begin{figure*}[t]
  \centering
  \includegraphics[width=\linewidth]{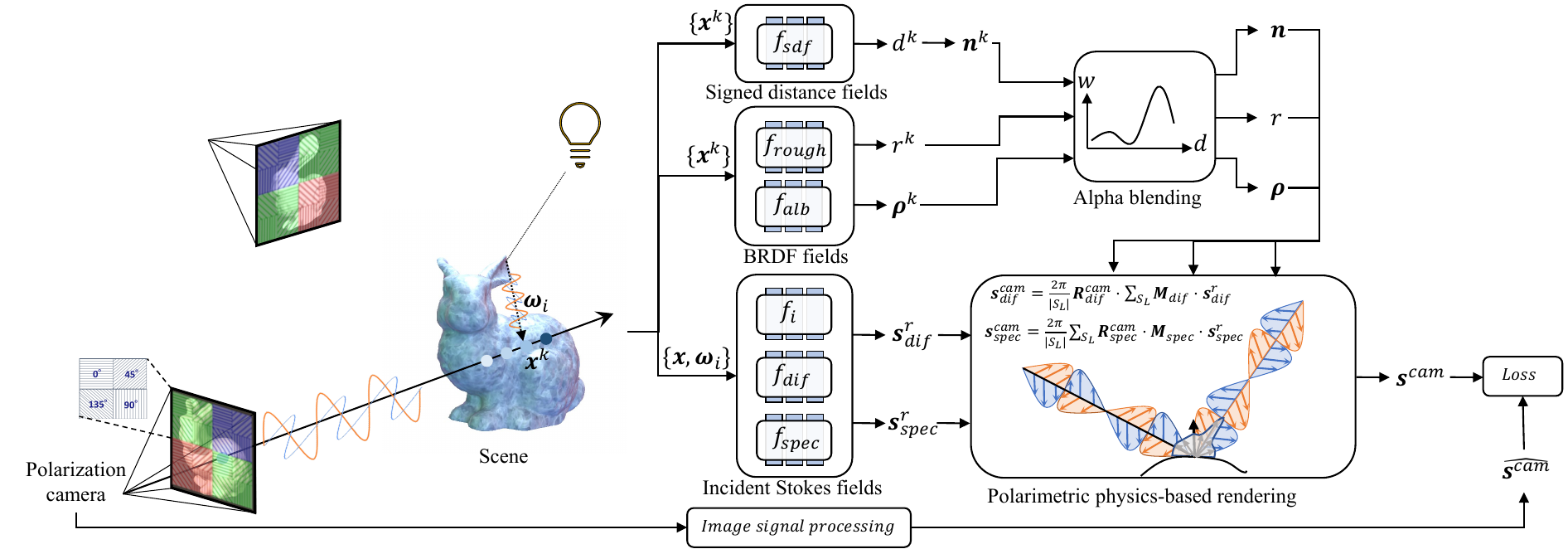}
  \caption{Overview of NeISF. For each interaction point, we use MLPs to implicitly record surface normal $\textbf{n}$ (Sec. \ref{sec:sdf fields}), diffuse albedo $\bm{\rho}$, roughness $r$ (Sec. \ref{sec:brdf fields}), and already-rotated incident Stokes vectors of diffuse $\textbf{s}^\text{r}_\text{dif}$ and specular $\textbf{s}^\text{r}_\text{spec}$ components (Sec. \ref{sec:incident stokes fields}). A physically-based polarimetric renderer is adopted to render the outgoing Stokes vectors $\textbf{s}^\text{cam}$ (Sec. \ref{sec:rendering}). }
  \label{fig:model}
\end{figure*}

\section{Our Approach}
Our method takes multi-view polarized images, masks, and camera poses as inputs and outputs diffuse albedo, roughness, and surface normal. It supports various downstream tasks including relighting, material editing, and diffuse-specular separation. Details will be introduced in the following subsections, and we show an overview of our method in Fig. \ref{fig:model}.
\subsection{Assumptions and scopes}
\label{sec:assumptions}
We keep the specular coefficient $\bm{k_s}=[1, 1, 1]$. In addition, we assume a constant refractive index $\eta=1.5$, because this is close to the refractive index of common materials such as acrylic glass (1.49), polypropylene plastic (1.49), and quartz (1.458). This work only focuses on object-level inverse rendering, and scene-level inverse rendering is beyond the scope. In addition, Baek pBRDF is only applicable to opaque and dielectric materials, which means objects that include metals, translucent, or transparent parts are not our target objects.

\subsection{Signed distance fields}
\label{sec:sdf fields}
We represent the geometry using a signed distance field net $f_\text{sdf}$. Let $\{ \mathbf{x}^k\}_{k=1}^N$ be the $N$ samples along the ray direction:
\begin{equation}
    f_\text{sdf}(\mathbf{x}^k) = d^k,
\end{equation}
$d^k$ is the signed distance from the nearest surface. The normal of the sampled location $\mathbf{x}^k$ can be obtained by calculating the normalized gradient of $f_\text{sdf}$:

\begin{equation}
    \nabla_{\mathbf{x}^k}f_\text{sdf}(\mathbf{x}^k) / ||\nabla_{\mathbf{x}^k}f_\text{sdf}(\mathbf{x}^k)||_2 = \mathbf{n}^k.
\end{equation}
After obtaining all the normals of sampled points, an alpha-blending is required to compute the surface normal of the interaction point. The weight $w^k$ of the alpha-blending can be calculated by:

\begin{equation}
    w^k = T^k(1-\exp{(-\sigma^k\delta^k)}),
\end{equation}
where $T^k=\exp{(-\sum_{j=1}^{k-1}\sigma^j\delta^j)}$, $\delta$ is the distance between two adjacent samples. For the density $\sigma$, we follow the definition of VolSDF \cite{yariv2021volume}:

\begin{equation}
    \sigma^k = \alpha\Psi_\beta(d^k),
\end{equation}
where $\Psi$ is the cumulative distribution function of the Laplace distribution, $\alpha$ and $\beta$ are two learnable parameters. 
Then, we compute the alpha-blending to achieve the final surface normal: $\mathbf{n} = \sum_{k=1}^N w^k \mathbf{n}^k$.

\subsection{BRDF fields}
\label{sec:brdf fields}
As the specular coefficient $\bm{k_s}$ and refractive index $\eta$ are assumed as constants, we only need to estimate the diffuse albedo $\bm{\rho}$ and roughness $r$. Thus, for each sampled location $\mathbf{x}^k$, we estimate:

\begin{equation}
    f_\text{alb}(\mathbf{x}^k) = \bm{\rho}^k,
\end{equation}

\begin{equation}
    f_\text{rough}(\mathbf{x}^k) = r^k.
\end{equation}
Similar to the surface normal, the albedo and roughness for the interaction point can also be calculated via alpha-blending: $\bm{\rho} = \sum_{k=1}^N w^k \bm{\rho}^k$, $r = \sum_{k=1}^N w^k r^k$.

%\begin{figure}[t]
%  \centering
%  \includegraphics[width=\linewidth]{implicit_explicit.pdf}
%  \caption{Two solutions of incident Stokes fields. Blue/orange light path represents the current procedure is implicit/explicit. (a) Modeling $\mathbf{s}^\text{in}$ implicitly, but $\mathbf{R}^\text{in}$ explicitly. (b) We model $\mathbf{R}^\text{in}$ implicitly as well. Thus, $\mathbf{s}^\text{r}$ is directly estimated using MLPS.}
%  \label{fig:implicit_explicit}
%\end{figure}

\begin{table*}[ht]
\begin{center}
\begin{tabular}{*7l}
\hline
{}  & {}& Ours & Ours-no-pol & PANDORA\cite{dave2022pandora} & NeILF++\cite{zhang2023neilf++} & VolSDF \cite{yariv2021volume}\\
\hline
\multirow{6}{*}{Bunny} &Normal (MAE$\downarrow$) & \textbf{1.727}$^{\circ}$ & $^{\hspace{1pc}}$3.295$^{\circ}$ & $^{\hspace{1pc}}$7.769$^{\circ}$ & $^{\hspace{0.9pc}}$4.481$^{\circ}$ & $^{\hspace{0.9pc}}$5.210$^{\circ}$ \\

                & Mixed (PSNR$\uparrow$) & \textbf{36.81} &$^{\hspace{1pc}}$34.45 &$^{\hspace*{1pc}}$24.62 & $^{\hspace{0.9pc}}$32.08 & $^{\hspace{0.9pc}}$35.08\\
                & Specular (PSNR$\uparrow$) & \textbf{27.68} &$^{\hspace{1pc}}$26.91 &$^{\hspace*{1pc}}$20.04 & $^{\hspace{1.72pc}}$- & $^{\hspace{1.72pc}}$-\\
                & Diffuse (PSNR$\uparrow$) & \textbf{35.31} &$^{\hspace{1pc}}$32.93 &$^{\hspace*{1pc}}$24.43 & $^{\hspace{1.72pc}}$- & $^{\hspace{1.72pc}}$-\\
                & Roughness (SI-L1$\downarrow$) & \textbf{.0149} &$^{\hspace{1pc}}$.0244 & $^{\hspace*{1.8pc}}$- & $^{\hspace*{1.72pc}}$- & $^{\hspace{1.72pc}}$- \\
                & Albedo (SI-L1$\downarrow$) & \textbf{.0372} &$^{\hspace{1pc}}$.0396 & $^{\hspace*{1.8pc}}$- & $^{\hspace*{1.72pc}}$- & $^{\hspace{1.72pc}}$- \\
                
\hline
\multirow{6}{*}{Teapot} &Normal (MAE$\downarrow$) & \textbf{2.541}$^{\circ}$ & $^{\hspace*{1pc}}$3.894$^{\circ}$ & $^{\hspace*{1pc}}$6.722$^{\circ}$ & $^{\hspace{0.9pc}}$5.752$^{\circ}$ & $^{\hspace{0.9pc}}$6.375$^{\circ}$ \\

                & Mixed (PSNR$\uparrow$) & \textbf{31.32} &$^{\hspace{1pc}}$30.23 &$^{\hspace{1pc}}$22.47 & $^{\hspace{0.9pc}}$25.24 & $^{\hspace{0.9pc}}$28.91\\
                & Specular (PSNR$\uparrow$) & \textbf{22.50} &$^{\hspace{1pc}}$21.15 &$^{\hspace{1pc}}$16.84 & $^{\hspace{1.72pc}}$- & $^{\hspace{1.72pc}}$-\\
                & Diffuse (PSNR$\uparrow$) & \textbf{31.97} &$^{\hspace{1pc}}$30.17 &$^{\hspace{1pc}}$22.39 & $^{\hspace{1.72pc}}$- & $^{\hspace{1.72pc}}$-\\
                & Roughness (SI-L1$\downarrow$) & \textbf{.0172} & $^{\hspace{1pc}}$.0223 &$^{\hspace{1.8pc}}$- &$^{\hspace{1.72pc}}$- &$^{\hspace{1.72pc}}$-\\
                & Albedo (SI-L1$\downarrow$) & \textbf{.0712} & $^{\hspace{1pc}}$.0720 &$^{\hspace{1.8pc}}$- &$^{\hspace{1.72pc}}$- &$^{\hspace{1.72pc}}$-\\
\hline
\end{tabular}
\end{center}

\caption{Results on synthetic dataset. Metrics are computed on 10 test images.  The surface normal is evaluated by mean angular error (MAE), and intensity images are evaluated with a peak signal-to-noise ratio (PSNR). Due to the inherent ambiguity of albedo and roughness, we use a scale-invariant L1 error (SI-L1) following IRON \cite{zhang2022iron}. "Mixed" represents the combination of "Specular" and "Diffuse".}
\label{table:compare_normal_rgb}
\end{table*}

\subsection{Incident Stokes fields}
\label{sec:incident stokes fields}
As NeILF \cite{yao2022neilf} proposed, the complicated multi-bounced light propagation can be represented as an incident light field. Specifically, given the location and direction of all second-last bounce lights, they use MLPs to record the light intensities. Seemingly, extending the incident light field to the incident Stokes vectors is straightforward, and the only thing we need to do is to change the outputs of MLPs from the 1D light intensities to the 3D Stokes vectors. However, as shown in Eq. \ref{equ:rendering equation dif} and Eq. \ref{equ:rendering equation spec}, rotation matrices must also be considered because Mueller-Stokes multiplication is only valid when they share the same reference frames. In addition, the diffuse and specular components have different behavior of rotations, which makes the problem even harder. One potential solution is to explicitly calculate rotation matrices $\mathbf{R}^\text{in}_\text{dif}$ and $\mathbf{R}^\text{in}_\text{spec}$. However, calculating the rotation matrices requires us to know the accurate reference frame of the current surface and incident light. The former can be easily calculated using the surface normal (for diffuse reflection) or half-vector (for specular reflection). However, computing the reference frame of the incident light is time consuming as it depends on the previous bounce, and explicitly simulating the previous bounce requires even more computational resources. Here we have an interesting observation: No matter what the reference frame of the incident light is, \textbf{what we care about is the value of the incident Stokes vectors after the rotation}. Thus, we propose a simple but efficient solution: modeling the rotation matrices implicitly. Specifically, instead of recording $\mathbf{s}^\text{in}$, we directly record the already-rotated Stokes vectors $\mathbf{R}^\text{in}_\text{dif}  \cdot \mathbf{s}^\text{in}$ and $\mathbf{R}^\text{in}_\text{spec}  \cdot \mathbf{s}^\text{in}$ using MLPs. For simplicity, we use $\mathbf{s}^\text{r}_\text{dif}$ and $\mathbf{s}^\text{r}_\text{spec}$ to denote the already-rotated incident Stokes vectors of diffuse and specular component separately.  Because the first elements (unpolarized light intensity) of $\mathbf{s}^\text{r}_\text{dif}$ and $\mathbf{s}^\text{r}_\text{spec}$ are the same, in practice, we use three MLPs to model the incident Stokes vectors. The first one is an incident intensity network:
\begin{equation}
    f_i(\mathbf{x}, \bm{\omega}_i) = \mathbf{s}^\text{r}_\text{spec} [0] = \mathbf{s}^\text{r}_\text{dif} [0], 
\end{equation}
where $[n]$ denotes the $n^\text{th}$ element of the vector. $\mathbf{x}$ is the ray-surface interaction point calculated using ray-marching. The second one is an incident specular Stokes network:
\begin{equation}
    f_\text{spec}(\mathbf{x}, \bm{\omega}_i) = \mathbf{s}^\text{r}_\text{spec} [1, 2], 
\end{equation}
and the third one is an incident diffuse Stokes network:
\begin{equation}
    f_\text{dif}(\mathbf{x}, \bm{\omega}_i) = \mathbf{s}^\text{r}_\text{dif} [1].
\end{equation}
Note that we do not estimate $\mathbf{s}^\text{r}_\text{dif} [2]$, as it will be canceled out in the polarimetric rendering. Please refer to the supplementary material for details.  

\subsection{Sphere sampling}
\label{sec:rendering}
Following NeILF \cite{yao2022neilf}, we solve the integral of the Rendering Equation using a fixed Fibonacci sphere sampling. So that we can rewrite Eq. \ref{equ:rendering equation dif} as follows:

\begin{equation}
    \mathbf{s}^\text{cam}_\text{dif} = \frac{2\pi}{|S_L|} \mathbf{R}^\text{cam}_\text{dif}  \cdot  \sum_{S_L} \mathbf{M}_\text{dif}  \cdot \mathbf{s}^\text{r}_\text{dif},
\end{equation}
where $\mathbf{s}^\text{cam}_\text{dif}$ is the outgoing Stokes vectors of the diffuse component, $S_L$ is the set of the sampled incident light over the hemisphere, $\mathbf{R}^\text{cam}_\text{dif}$ is the rotation matrix computed using the estimated surface normal, $\mathbf{M}_\text{dif}$ is the estimated Mueller matrix of the diffuse component, and $\mathbf{s}^\text{r}_\text{dif}$ is the incident diffuse Stokes vectors. Similarly, we can also rewrite Eq. \ref{equ:rendering equation spec} for the specular component:
\begin{equation}
    \mathbf{s}^\text{cam}_\text{spec} = \frac{2\pi}{|S_L|}  \sum_{S_L} \mathbf{R}^\text{cam}_\text{spec} \cdot  \mathbf{M}_\text{spec}  \cdot \mathbf{s}^\text{r}_\text{spec}.
\end{equation}
The final output can be obtained by:
\begin{equation}
    \mathbf{s}^\text{cam} = \mathbf{s}^\text{cam}_\text{dif} + \mathbf{s}^\text{cam}_\text{spec}.
\end{equation}

\subsection{Training scheme}
\label{sec:training}
We use a three-stage training scheme. The first stage initializes the geometry. Specifically, we train VolSDF \cite{yariv2021volume} to learn a signed distance field $f_\text{sdf}$. The second stage initializes the material and lighting. And, this stage does not update the signed distance field $f_\text{sdf}$. The other neural fields are optimized with the $L_1$ loss on the estimated Stokes vectors $\mathbf{s}^\text{cam}$ and their ground truth $\hat{\mathbf{s}^\text{cam}}$. In the third stage, we jointly optimize all the neural fields. In addition to the $L_1$ loss, we also compute an Eikonal loss $L_\text{Eik}$ \cite{yariv2021volume} to regularize the signed distance field $f_\text{sdf}$.

% losses
\section{Experiments}
\subsection{Datasets}
We introduce one synthetic dataset and one real-world dataset for the model evaluation. Although PANDORA \cite{dave2022pandora} also proposes a synthetic polarimetric dataset, the scene setup is simpler than common real-world scenarios. Specifically, the object is illuminated by an unpolarized environment map such that almost all incident light is unpolarized. To solve this problem, we place the object inside an altered ``Cornell Box" to mimic real-world situations, where the light is bounced multiple times and becomes polarized before interacting with the object. For each object, we render 110 HDR polarized images using Mitsuba 3.0 \cite{Jakob2020DrJit} with Baek pBRDF. Among them, 100 images are used for training and 10 are used for testing. We also capture a real-world HDR dataset, as most existing polarimetric datasets are LDR, which may affect the training due to saturation and unknown gamma correction. We capture the polarized images using a polarization camera (FLIR BFS-U3-51S5PC-C). For each viewpoint, we capture images with different exposure times and composite them to obtain one HDR image. We selected three real-world objects, and for each object, we captured 96 views for training and 5 views for evaluation. We recommend our readers check the supplementary documents for details.

%\begin{table}[ht]
%\begin{center}
%\setlength\tabcolsep{2pt}
%\begin{tabular}{*4c}
%\hline
%{}  & PANDORA \cite{dave2022pandora}& NeILF++\cite{zhang2023neilf++} & Ours\\
%\hline

%Polarization cues    & \cmark & \xmark & \cmark\\
%Multi-bounce    & \xmark & \cmark & \cmark\\
%BRDF estimation & \xmark & \cmark & \cmark\\
%\hline
%\end{tabular}
%\end{center}
%\caption{Differences between our method and the competitors.}
%\label{table:compare_methods}
%\end{table}

%\begin{table}[ht]
%\begin{center}
%\begin{tabular}{*4l}
%\hline
%{}  & {}& Ours & Ours-no-pol\\
%\hline
%\multirow{2}{*}{Bunny} 

%                & Roughness (SI-L1$\downarrow$) & \textbf{.0149} &.0244 \\
%                & Albedo (SI-L1$\downarrow$) & \textbf{.0372} &.0396 \\
%\hline
%\multirow{2}{*}{Teapot} 

%                & Roughness (SI-L1$\downarrow$) & \textbf{.0174} & .0223\\
%                & Albedo (SI-L1$\downarrow$) & \textbf{.0716} & .0720\\
%\hline
%\end{tabular}
%\end{center}
%\caption{BRDF parameters estimation results of NeISF-synthetic. For each object, metrics are computed on the average of 10 test images. Due to the inherent ambiguity of albedo and roughness, we use a scale-invariant L1 loss following IRON \cite{zhang2022iron}.}
%\label{table:compare_brdf}
%\end{table}

\begin{figure*}[ht]
  \centering
  \includegraphics[width=\linewidth]{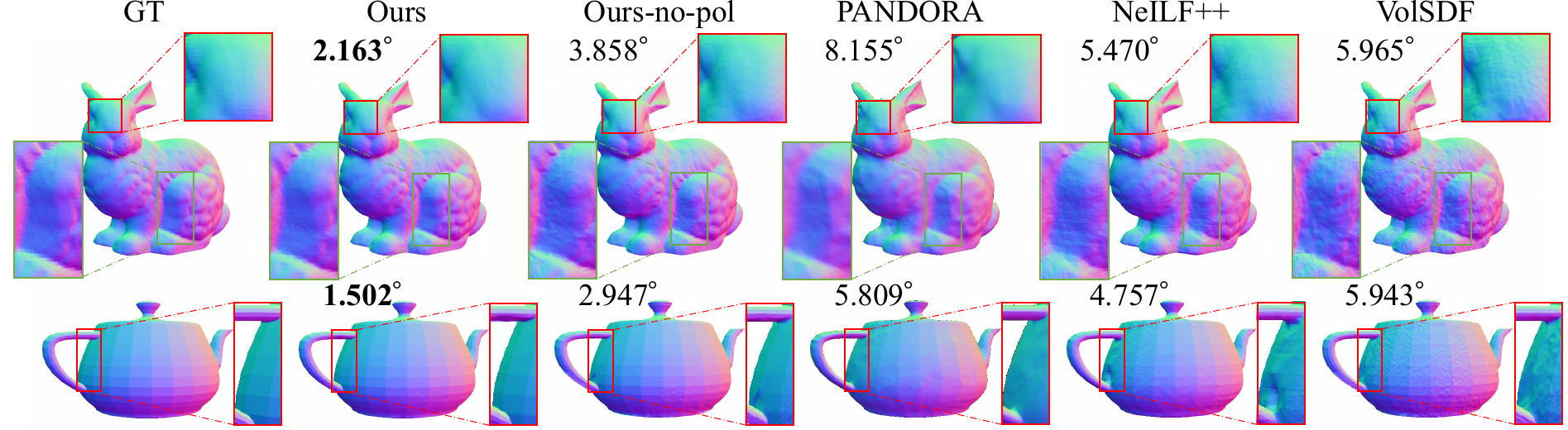}
  \caption{The reconstructed surface normal on the synthetic dataset. MAE metrics are on the top left. PANDORA \cite{dave2022pandora} loses details of surface normals.  NeILF++ \cite{zhang2023neilf++}, VolSDF \cite{yariv2021volume}, and ours-no-pol fail to disentangle the geometry and material. Thus, the estimated surface normals contain some patterns that come from the albedo.}
  \label{fig:compare_normal_synthetic}
\end{figure*}

\begin{figure}[ht]
  \centering
  \includegraphics[width=\linewidth]{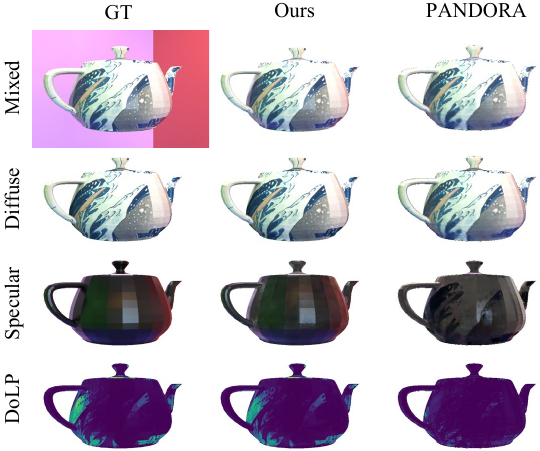}
  \caption{Diffuse-specular separation and DoLP images of synthetic dataset. We can observe the reflection of green and red walls on the teapot for our method, where PANDORA \cite{dave2022pandora} fails.}
  \label{fig:compare_diffuse_specular_teapot}
\end{figure}

\begin{figure}[ht]
  \centering
  \includegraphics[width=\linewidth]{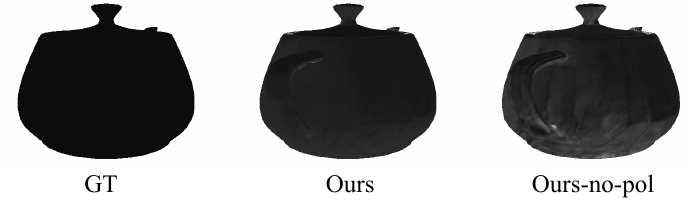}
  \caption{Roughness comparison. Without polarization cues, the recovered roughness is easily affected by geometry and shadows.}
  \label{fig:compare_roughness_teapot}
\end{figure}

\begin{figure*}[ht]
  \centering
  \includegraphics[width=0.985\linewidth]{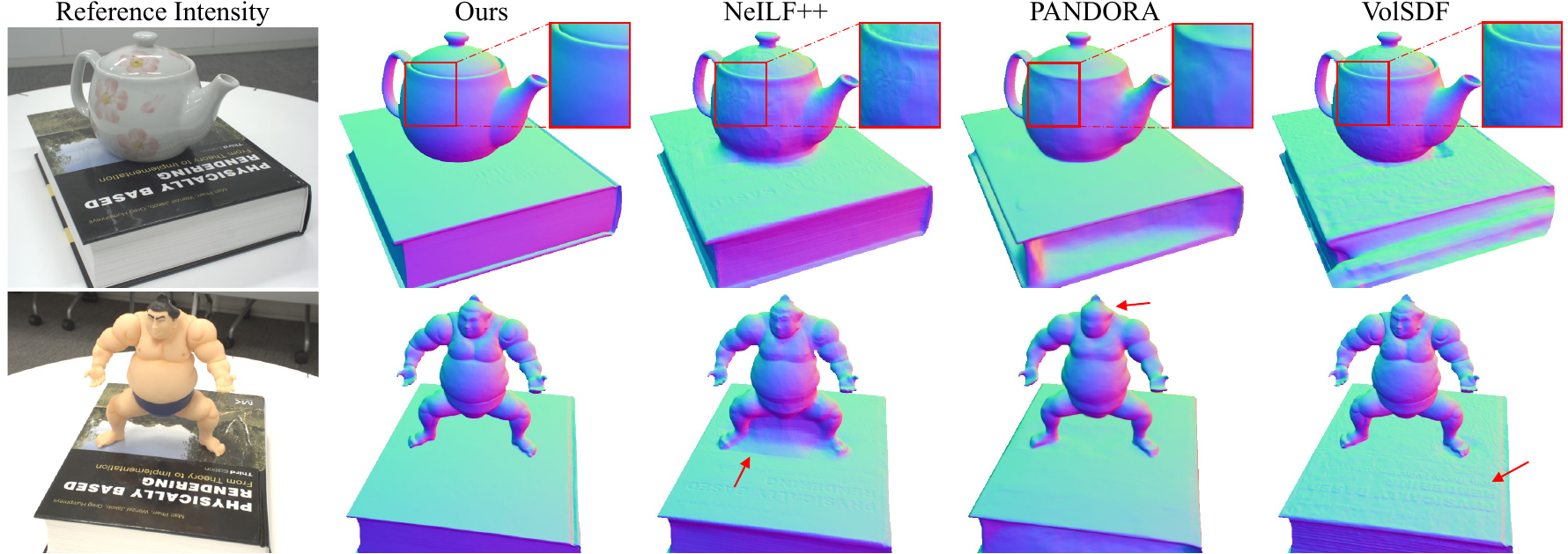}
  \caption{Qualitative comparison on real dataset. NeILF++ \cite{zhang2023neilf++}, PANDORA \cite{dave2022pandora}, and VolSDF \cite{yariv2021volume} misinterpret materials as geometries. For example the flower pattern on the teapot and the text on the surface of the book. However, this is not correct because these patterns come from the albedo. On the other hand, our method can reconstruct a clean surface normal.}
  \label{fig:compare_real_normal}
\end{figure*}

\begin{figure}[ht]
  \centering
  \includegraphics[width=\linewidth]{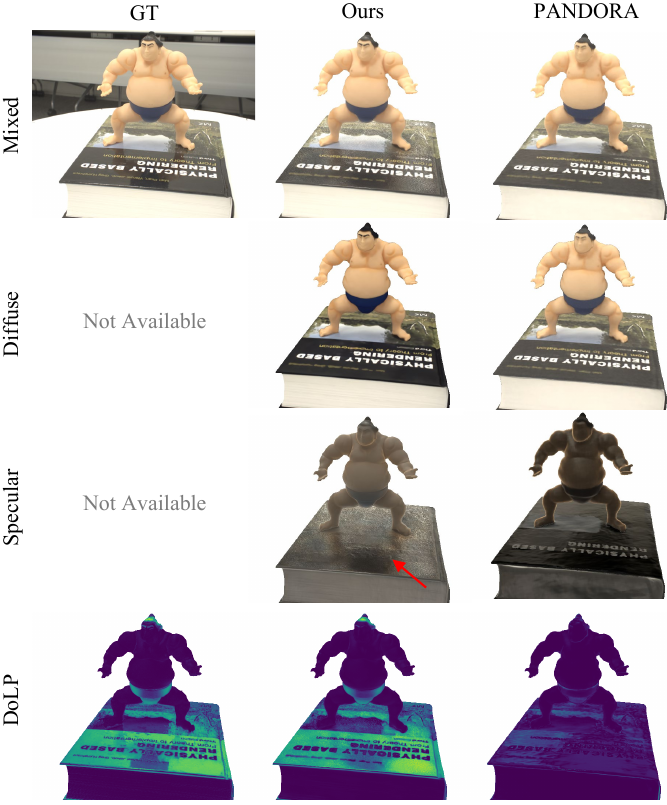}
  \caption{Diffuse-specular separation and DoLP images of real dataset. In the specular image, we can clearly observe the reflection on the surface of the book, but PANDORA \cite{dave2022pandora} fails to reconstruct such kinds of results due to the single-bounce assumption. In addition, our DoLP is visually similar to the GT.}
  \label{fig:compare_diffuse_specular_sumo}
\end{figure}

\begin{figure}[ht]
  \centering
  \includegraphics[width=\linewidth]{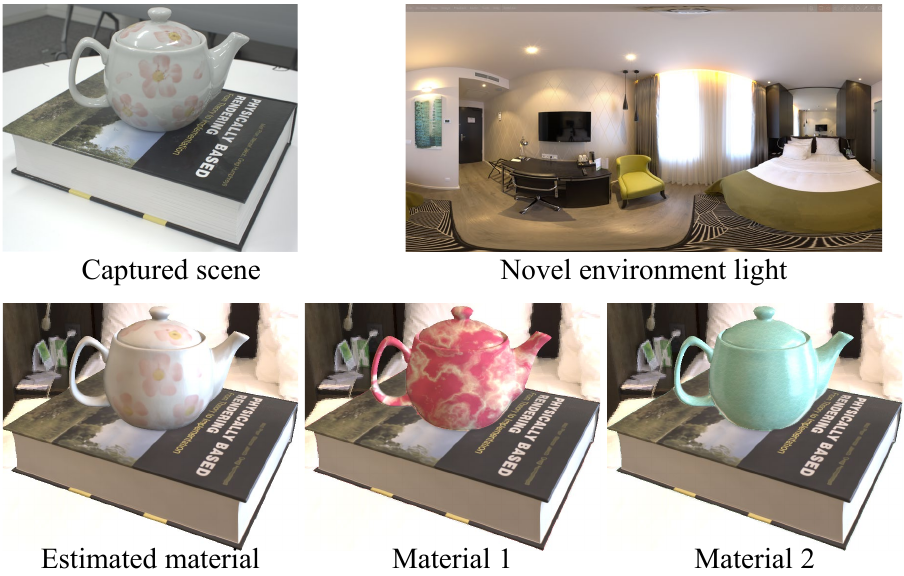}
  \caption{Relighting and material editing results. We edit the material of the teapot. Due to the accurate disentanglement of geometry and material, the edited image has realistic reflections.}
  \label{fig:relighting_material_editing}
\end{figure}

\subsection{Baselines}
Looking for competitors for our proposal is not easy. Most of the NeRF-based inverse rendering works \cite{boss2021nerd, zhang2022modeling, munkberg2022extracting, yao2022neilf, zhang2023neilf++, hasselgren2022shape, jin2023tensoir, wu2023nefii, cheng2023wildlight, liu2023nero} use Disney BRDF \cite{burley2012physically} model for rendering. Although they also estimate parameters such as roughness and albedo, these parameters have different physical meanings from ours, as Baek pBRDF is not based on Disney BRDF \cite{burley2012physically}. Nevertheless, the estimated surface normal as well as the reconstructed intensity images can be compared. Thus, we chose \textbf{VolSDF} \cite{yariv2021volume} and the latest NeRF-based inverse rendering work \textbf{NeILF++} \cite{zhang2023neilf++} as our competitors. Because the official implementation of VolSDF does not support HDR images as inputs, we use our own implementation. Besides, \textbf{PANDORA} \cite{dave2022pandora} is also considered as a baseline method. Although they do not support estimating BRDF parameters, the surface normal, diffuse-specular separation, and reconstructed polarized images are comparable. An important ablation study should be the performance with or without the presence of polarization cues. To achieve this, we introduce an unpolarized version of NeISF. Specifically, we remove $f_\text{dif}$ and $f_\text{spec}$ and only keep $f_i$. In addition, we also implement an unpolarized version of Beak pBRDF for rendering. Finally, the loss is only computed on the intensity space. The other parts are exactly the same as our model. We denote this model as \textbf{Ours-no-pol}, and it can also be considered as a variant of NeILF \cite{yao2022neilf}. Details of this unpolarized BRDF can be found in the supplementary material.

\subsection{Results}
\noindent \textbf{Synthetic Dataset} We report the quantitative results of the surface normal, intensity, diffuse-specular separation, roughness, and albedo in Tab. \ref{table:compare_normal_rgb}. VolSDF does not support diffuse-specular separation. Although NeILF++ supports diffuse-specular separation, the diffuse and specular images differ from our physical meanings. Thus, we do not report diffuse-specular separation for these two methods. For the qualitative comparison, we show the surface normal results in Fig. \ref{fig:compare_normal_synthetic}, the diffuse-specular separation, and DoLP results in Fig. \ref{fig:compare_diffuse_specular_teapot}, and the roughness results in Fig. \ref{fig:compare_roughness_teapot}.

\noindent\textbf{Real Dataset} We show the surface normal results in Fig. \ref{fig:compare_real_normal}, the diffuse-specular separation, and DoLP results in Fig. \ref{fig:compare_diffuse_specular_sumo}, and the material editing and relighting results in Fig. \ref{fig:relighting_material_editing}.  Detailed analysis can be found in the figure captions.

\section{Limitations}
\label{sec:limitation}
Several limitations still exist. First, the implicit Stokes representation is a double-edged sword. It allows us to model complicated polarimetric light transportation. At the same time, the estimated lighting can not be used in the conventional renderer. Second, the current solution only considers opaque dielectric objects. However, polarization cues can also provide rich information for translucent or metal objects. Finally, due to the manufacturing design of the polarization sensors, the captured four polarized images are not perfectly aligned. This makes the Stokes vectors of real-world data so noisy, making it impossible to handle high-frequency signals such as small bumps or edges.

\section{Conclusion}
We have proposed NeISF, an inverse rendering pipeline that combines implicit scene representations and polarization cues. It relies on the following novelties. The first one is an implicit representation of the multi-bounced Stokes vectors which takes care of the rotations. The second one is a physically-based polarimetric renderer. With these two novelties, NeISF outperforms the existing inverse render models for both synthetic and real-world datasets. The ablation study has verified the contribution of polarization cues. However, several limitations mentioned in Sec. \ref{sec:limitation} still exist and are worth further exploration.

%%%%%%%%% REFERENCES
{\small
\bibliographystyle{ieee_fullname}
\bibliography{egbib}
}

\end{document}